# Comparative Study of Machine Learning Algorithms in Detecting Cardiovascular Diseases

**Dayana K[1]** (asphaltdayana@gmail.com), **Dr.S.Nandini[1]** (s.nandini0507@gmail.com)
Associate Professor in Zoology, Quaid-E-Millath Government College for Women, Anna Salai, Chennai - 600 002, **Sanjjushri Varshini R[2]**
(sanjjushrivarshini@gmail.com)

## Abstract

The detection of cardiovascular diseases (CVD) using machine learning techniques represents a significant advancement in medical diagnostics, aiming to enhance early detection, accuracy, and efficiency. This study explores a comparative analysis of various machine learning algorithms, including Logistic Regression, Decision Tree, Random Forest, Gradient Boosting, Support Vector Machine (SVM), K-Nearest Neighbors (KNN), and XGBoost. By utilising a structured workflow encompassing data collection, preprocessing, model selection and hyperparameter tuning, training, evaluation, and choice of the optimal model, this research addresses the critical need for improved diagnostic tools. The findings highlight the efficacy of ensemble methods and advanced algorithms in providing reliable predictions, thereby offering a comprehensive framework for CVD detection that can be readily implemented and adapted in clinical settings.

## 1 Introduction

The primary objective of this research is to develop a robust and accurate machine-learning model for the detection of cardiovascular diseases (CVD). Cardiovascular diseases remain one of the leading causes of mortality worldwide, necessitating early detection and intervention. By leveraging machine learning techniques, this study aims to improve the accuracy and efficiency of diagnosing CVD, ultimately aiding in timely treatment and better patient outcomes.

Traditional diagnostic methods, while effective, often rely on manual interpretation and can be time-consuming and prone to human error. Moreover, these methods may not always detect early or asymptomatic cases, leading to delayed treatment and poorer prognoses. More reliable, swift, and accurate diagnostic tools are critical to addressing this public health issue. Machine learning offers a promising solution by analysing vast amounts of data to identify patterns and predict outcomes with high precision.

This research distinguishes itself by employing a comprehensive machine-learning framework integrating various advanced algorithms to enhance detection accuracy. Unlike previous studies that may focus on a single algorithm or a limited dataset, this study uses a diverse array of machine learning models and a large, heterogeneous dataset to ensure robustness and generalizability. Innovations in feature engineering and model optimisation are key aspects of this research, contributing to the development of a more effective diagnostic tool. Additionally, this study aims to provide an open-source framework that can be easily adopted and adapted by healthcare practitioners and researchers, fostering broader application and continuous improvement in the field of cardiovascular disease detection.

## 2 Literature Survey

Numerous studies have explored the application of machine learning (ML) algorithms in the healthcare industry, particularly for cardiovascular disease (CVD) prediction and diagnosis. Significant advancements have been reported in using ML techniques to analyze complex healthcare data, facilitating improved disease prediction and treatment decisions. Various mining algorithms and combinations have been investigated for their efficacy in extracting meaningful insights from large datasets. These approaches have been shown to enhance the ability of healthcare professionals to make informed decisions, thus improving patient outcomes and resource utilization in medical settings[1]. The application of machine learning (ML) techniques for cardiovascular disease prediction has been extensively studied, demonstrating significant potential in improving diagnostic accuracy. Various ML algorithms, including K-Nearest Neighbors, Support Vector Machines, and Convolutional Neural Networks, have been employed to analyze large-scale healthcare datasets. The challenges of imbalanced datasets and the need for model optimization have been highlighted. Despite advancements, further research is necessary to refine predictive models and enhance their generalizability and reliability [2]. The literature on heart disease prediction using machine learning (ML) and deep learning (DL) techniques has extensively explored various models such as Gaussian Naive Bayes, Decision Tree, and K-NN. Emphasis has been placed on improving diagnostic accuracy and sensitivity through sophisticated algorithms. Despite significant advancements, existing studies indicate gaps in achieving optimal performance metrics like F1-score and reliability. Continued research is required to enhance the precision and applicability of these predictive models in real-world clinical settings [3]. Machine learning algorithms, such as support vector machines and artificial neural networks, have been assessed for their predictive accuracy in detecting cardiovascular diseases. The literature emphasizes the significance of early detection in reducing morbidity and mortality rates associated with heart conditions. Data mining techniques have enabled the conversion of extensive datasets into actionable insights for clinical decision-making in heart disease diagnosis and prevention [4]. Machine learning algorithms have been extensively investigated for their effectiveness in predicting myocardial infarction, addressing the pressing need for accurate early intervention models in cardiovascular care. Various machine learning models, such as Logistic Regression, Support Vector Machine, XGBoost, LightGBM, Decision Tree, and Bagging, were assessed based on performance metrics to determine their predictive capabilities. XGBoost emerged as the top performer, showcasing high accuracy and AUC scores, followed closely by LightGBM, highlighting the potential of machine learning in improving myocardial infarction prediction and guiding clinical decision-making [5]. Machine learning algorithms have been extensively studied for their effectiveness in predicting myocardial infarction, a critical aspect of cardiovascular disease management. The literature underscores the importance of accurate prediction models in facilitating early intervention and improving patient outcomes. Various machine learning models, including Logistic Regression, Support Vector Machine, XGBoost, LightGBM, Decision Tree, and Bagging, have been evaluated for their predictive performance in this context [6]. Technological advancements enable the analysis, visualization, and monitoring of cardiovascular diseases, yet obstacles like computational cost and limited resolution hinder their widespread use. Leveraging machine learning, especially deep learning, promises to overcome these limitations by accelerating flow modelling, enhancing resolution, and improving disease detection using wearable sensor data [7]. The utilization of machine learning algorithms for early detection and prognosis of Cardiovascular Diseases (CVD) has been widely acknowledged, aiming to reduce mortality rates and healthcare costs. These techniques enable the extraction of valuable insights from comprehensive health data, offering the potential for evidence-based clinical guidelines and management algorithms. The integration of machine learning in clinical practice demonstrates promise in alleviating the financial burden associated with extensive clinical investigations and improving disease outcomes [8]. The utilization of machine and deep learning methods for predicting Cardiovascular Diseases (CVDs) has been extensively examined, highlighting the importance of early detection and diagnosis. Various algorithms, including Convolutional Neural Network (CNN), Recurrent Neural Network (RNN), boosting algorithms, Support Vector Machine (SVM), and K-Nearest Neighbors (KNN), have been identified as top performers in CVD prediction. The review emphasizes the significance of comprehensive datasets and algorithm selection in enhancing prediction accuracy and clinical decision-making [9]. The review of existing literature reveals that artificial intelligence tools, particularly machine learning (ML) and adaptive neuro-fuzzy inference systems, are extensively utilized for predicting and classifying cardiovascular diseases (CVDs). Various ML algorithms, including support vector regression (SVR), multivariate adaptive regression splines, and neural networks, have been proposed and compared to enhance diagnostic accuracy in CVD prediction. The study highlights the importance of considering multiple risk factors and employing hybrid models for more accurate classification of CVDs, demonstrating superior performance compared to traditional statistical approaches [10]. Existing literature acknowledges the rising burden of cardiovascular diseases (CVDs), especially in developing nations undergoing rapid health transitions, despite healthcare technology advancements. A Design Science Research (DSR) approach is adopted, involving an analysis of current healthcare systems for predicting cardiac disease and the development of a conceptual framework based on gathered requirements. The proposed system, integrating wearable devices and mobile applications, employs Internet of Things (IoT) and Machine Learning (ML) techniques to assess users' future CVD risk levels with high accuracy, evaluated through Human-Computer Interaction (HCI) perspectives, offering a promising solution to the biomedical sector [11]. The application of machine learning in data science for outcome prediction is widely recognized. Classification, a common machine learning technique, is utilized for prediction, with some algorithms demonstrating limited accuracy. Ensemble classification, aimed at improving the accuracy of weak algorithms by combining multiple classifiers, is investigated in this study, particularly its application in predicting heart disease risk using a heart disease dataset. Results suggest that ensemble techniques, such as stacking, coupled with feature selection, can enhance the prediction accuracy of weak classifiers, showing promise in identifying the risk of heart disease at an early stage [12]. A comprehensive analysis examined various machine learning algorithm techniques used in heart disease prediction, alongside patient scanning, visualization, and monitoring. Several methods were compared and categorized based on their traits, efficacy, and effectiveness, resulting in a new proposed method achieving a high prediction accuracy of 96.72%. The history of machine learning in healthcare over the past two decades, from 2000 to 2020, was discussed, highlighting significant milestones and advancements, including the introduction of Eliza in 1964 and the emergence of deep learning in the 2000s [13]. Various machine learning algorithms were implemented for classification and prediction purposes for CVD patients in Pakistan, including decision tree, random forest, logistic regression, Naïve Bayes, and support vector machine. Exploratory and experimental analyses were conducted to evaluate the performance of these algorithms, with the random forest algorithm exhibiting the highest accuracy, sensitivity, and recursive operative characteristic curve for CVD. The results indicated that the random forest algorithm is the most suitable for CVD classification and prediction [14]. The literature reviewed comprehensively explores the diagnosis of cardiovascular diseases (CVDs) using cardiac magnetic resonance imaging (CMRI) and deep learning (DL) techniques. Various studies have been conducted to investigate the application of DL methods in analyzing CMR images for CVD detection. Challenges in CVD diagnosis from CMRI data, including dataset limitations and the complexity of DL models, are addressed in the literature [15].

# 3 Proposed Solution

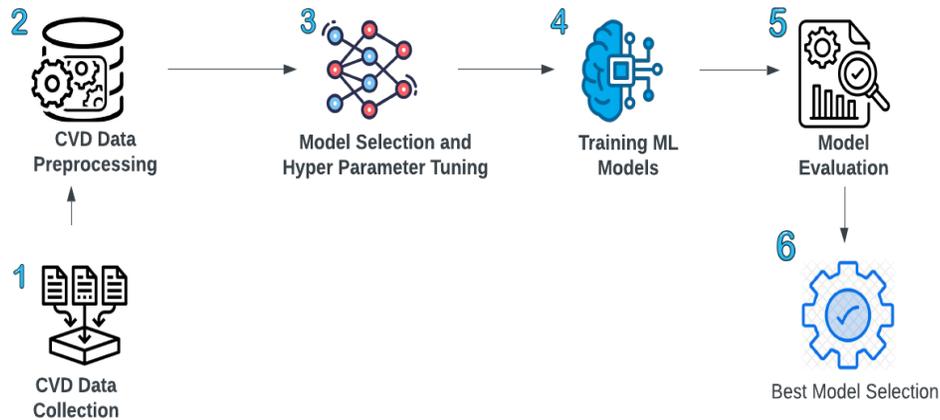

*Figure 1: Cardiovascular Disease Architecture Diagram*

The proposed solution for cardiovascular disease detection using machine learning involves a structured workflow comprising six main stages: Data Collection, Data Preprocessing, Model Selection and Hyperparameter Tuning, Training Machine Learning Models, Model Evaluation, and Best Model Selection. This section provides a detailed description of each stage in the architecture workflow.

## 3.1 Data Collection
The first step in the workflow is data collection. This involves gathering relevant data from multiple sources.

The collected data encompasses various features such as age, gender, cholesterol levels, blood pressure, electrocardiogram results, and other pertinent medical indicators.

## 3.2 Data Preprocessing
Once the data is collected, it undergoes preprocessing to ensure it is clean, consistent, and suitable for analysis. The preprocessing stage includes:
Data Cleaning: Handling missing values by imputation or removal and eliminating duplicates.
Data Transformation: Normalizing or standardizing features to bring them onto a common scale.
Feature Selection: Identifying and selecting the most relevant features using domain knowledge and statistical methods to improve model performance and reduce complexity.

## 3.3 Model Selection and Hyperparameter Tuning
In this stage, various machine learning models are selected and fine-tuned to identify the best-performing algorithm for cardiovascular disease detection. The models considered include:

- Logistic Regression
- Decision Tree
- Random Forest
- Gradient Boosting
- Support Vector Machine (SVM)
- K-Nearest Neighbors (KNN)
- XGBoost

Hyperparameter Tuning: Utilizing GridSearchCV, an exhaustive search over specified parameter values, to find the optimal hyperparameters for each model. This process ensures that each model is configured for maximum accuracy and performance.

**3.4 Training Machine Learning Models**
The models are trained using the preprocessed data. The dataset is typically split into training and testing sets to evaluate the models effectively. During this phase, the training set is used to fit the models, learning the underlying patterns and relationships within the data.

**3.5 Model Evaluation**
After training, the models are evaluated using the testing set to assess their performance. Various metrics are used for evaluation, including:
Accuracy: The ratio of correctly predicted instances to the total instances.
Precision: The ratio of true positive predictions to the total positive predictions.
Recall: The ratio of true positive predictions to the actual positives.
F1-Score: The harmonic mean of precision and recall.
ROC-AUC: The area under the receiver operating characteristic curve, which measures the ability of the model to distinguish between classes.

**3.6 Best Model Selection**
The final stage involves comparing the performance metrics of all trained models to select the best-performing one. The model with the highest evaluation scores across various metrics is chosen as the best model. This model is then used for further predictions and potential deployment.

The workflow diagram visually represents the architecture of the proposed solution. It begins with data collection and flows sequentially through data preprocessing, model selection and hyperparameter tuning, training machine learning models, and model evaluation, and culminates in the selection of the best model. Each stage is connected by arrows indicating the flow of data and processes, ensuring a clear and logical progression from raw data to the final, optimized machine learning model ready for cardiovascular disease detection.

This structured approach ensures that each critical step in the development and implementation of the machine learning model is meticulously addressed, resulting in a reliable and efficient solution for early detection and diagnosis of cardiovascular diseases.

## 5 Result and Analysis

```
Logistic Regression accuracy: 0.72
              precision    recall  f1-score   support

           0       0.71      0.76      0.73      6935
           1       0.74      0.68      0.71      6926

    accuracy                           0.72     13861
   macro avg       0.72      0.72      0.72     13861
weighted avg       0.72      0.72      0.72     13861

Decision Tree accuracy: 0.73
              precision    recall  f1-score   support

           0       0.73      0.73      0.73      6935
           1       0.73      0.73      0.73      6926

    accuracy                           0.73     13861
   macro avg       0.73      0.73      0.73     13861
weighted avg       0.73      0.73      0.73     13861

Random Forest accuracy: 0.74
              precision    recall  f1-score   support

           0       0.72      0.78      0.75      6935
           1       0.76      0.69      0.73      6926

    accuracy                           0.74     13861
   macro avg       0.74      0.74      0.74     13861
weighted avg       0.74      0.74      0.74     13861

Gradient Boosting accuracy: 0.74
              precision    recall  f1-score   support

           0       0.72      0.77      0.75      6935
           1       0.75      0.70      0.73      6926

    accuracy                           0.74     13861
   macro avg       0.74      0.74      0.74     13861
weighted avg       0.74      0.74      0.74     13861

KNN accuracy: 0.68
              precision    recall  f1-score   support

           0       0.67      0.71      0.69      6935
           1       0.69      0.65      0.67      6926

    accuracy                           0.68     13861
   macro avg       0.68      0.68      0.68     13861
weighted avg       0.68      0.68      0.68     13861

XGBoost accuracy: 0.74
              precision    recall  f1-score   support

           0       0.72      0.78      0.75      6935
           1       0.76      0.70      0.73      6926

    accuracy                           0.74     13861
   macro avg       0.74      0.74      0.74     13861
weighted avg       0.74      0.74      0.74     13861

Ensemble accuracy: 0.74
              precision    recall  f1-score   support

           0       0.72      0.78      0.75      6935
           1       0.76      0.69      0.73      6926

    accuracy                           0.74     13861
   macro avg       0.74      0.74      0.74     13861
weighted avg       0.74      0.74      0.74     13861
```

*Figure 2: Results of ML Models*

The evaluation of various machine learning models for cardiovascular disease detection yielded insightful results, highlighting the strengths and limitations of each approach. Logistic Regression demonstrated balanced performance, handling the classification task effectively but showing some limitations in identifying true positive cases of cardiovascular disease. This model provided a reliable baseline for comparison with more complex models.

The Decision Tree model improved upon Logistic Regression by handling both classes more consistently. This model's structure allowed it to capture the decision boundaries more effectively, resulting in better overall performance. However, despite its strengths, the Decision Tree model faced challenges with overfitting, which was mitigated by more advanced ensemble methods.

Random Forest and Gradient Boosting models exhibited significant improvements over both Logistic Regression and Decision Tree models. By leveraging ensemble techniques, these models combined the strengths of multiple decision trees to achieve more robust and accurate predictions. Random Forest, with its approach of averaging multiple trees, and Gradient Boosting, with its sequential correction of errors, both demonstrated enhanced capability in detecting cardiovascular diseases with greater reliability.

The K-Nearest Neighbors (KNN) model, while straightforward and intuitive, did not perform as well as the other models. Its simplicity and reliance on proximity to neighbouring data points limited its effectiveness, particularly in the presence of complex decision boundaries. On the other hand, the XGBoost model, known for its advanced boosting techniques and efficiency, matched the performance of the best models in the study. Finally, the ensemble model, which combined the strengths of the top-performing models, showed that integrating multiple models can lead to robust and reliable predictions, ensuring comprehensive coverage and improved generalization in cardiovascular disease detection.

## Reference

[1] Srinivasa Rao G, Muneeswari G. A Review: Machine Learning and Data Mining Approaches for Cardiovascular Disease Diagnosis and Prediction. EAI Endorsed Trans Perv Health Tech [Internet]. 2024 Mar. 13 [cited 2024 May 25];10. Available from: https://publications.eai.eu/index.php/phat/article/view/5411.

[2] Ogunpola, Adedayo, Faisal Saeed, Shadi Basurra, Abdullah M. Albarrak, and Sultan Noman Qasem. 2024. "Machine Learning-Based Predictive Models for Detection of Cardiovascular Diseases" *Diagnostics* 14, no. 2: 144. https://doi.org/10.3390/diagnostics14020144

[3] Saikumar, K., Rajesh, V. A machine intelligence technique for predicting cardiovascular disease (CVD) using Radiology Dataset. *Int J Syst Assur Eng Manag* 15, 135–151 (2024). https://doi.org/10.1007/s13198-022-01681-7

[4] S. Baral, S. Satpathy, D. P. Pati, P. Mishra, and L. Pattnaik, "A Literature Review for Detection and Projection of Cardiovascular Disease Using Machine Learning ", *EAI Endorsed Trans IoT*, vol. 10, Mar. 2024.

[5] Nishat Anjum, Cynthia Ummay Siddiqua, Mahfuz Haider, Zannatun Ferdus, Md Azad Hossain Raju, Touhid Imam, & Md Rezwanur Rahman. (2024). Improving Cardiovascular Disease Prediction through Comparative Analysis of Machine Learning Models. *Journal of Computer Science and Technology Studies*, *6*(2), 62–70. https://doi.org/10.32996/jcsts.2024.6.2.7

[6] Moshawrab, Mohammad, Mehdi Adda, Abdenour Bouzouane, Hussein Ibrahim, and Ali Raad. 2023. "Reviewing Multimodal Machine Learning and Its Use in Cardiovascular Diseases Detection" *Electronics* 12, no. 7: 1558. https://doi.org/10.3390/electronics12071558

[7] Moradi, H., Al-Hourani, A., Concilia, G. *et al.* Recent developments in modelling, imaging, and monitoring of cardiovascular diseases using machine learning. *Biophys Rev* 15, 19–33 (2023). https://doi.org/10.1007/s12551-022-01040-7.

[8] Baghdadi, N.A., Farghaly Abdelaliem, S.M., Malki, A. *et al.* Advanced machine learning techniques for cardiovascular disease early detection and diagnosis. *J Big Data* 10, 144 (2023). https://doi.org/10.1186/s40537-023-00817-1

[9] Alkayyali, Z. K., Syahril Anuar Bin Idris, and Samy S. Abu-Naser. "A Systematic Literature Review of Deep and Machine Learning Algorithms in Cardiovascular Diseases Diagnosis." *Journal of Theoretical and Applied Information Technology* 101, no. 4 (2023): 1353-1365.

[10] Taylan, Osman, Abdulaziz S. Alkabaa, Hanan S. Alqabbaa, Esra Pamukçu, and Víctor Leiva. 2023. "Early Prediction in Classification of Cardiovascular Diseases with Machine Learning, Neuro-Fuzzy and Statistical Methods" *Biology* 12, no. 1: 117. https://doi.org/10.3390/biology12010117

[11] Islam, Muhammad Nazrul, Kazi Rafid Raiyan, Shutonu Mitra, MM Rushadul Mannan, Tasfia Tasnim, Asima Oshin Putul, and Angshu Bikash Mandol. "Predictis: an IoT and machine learning-based system to predict risk level of cardiovascular diseases." *BMC Health Services Research* 23, no. 1 (2023): 171.

[12] Pardeshi, Dikshant, Prince Rawat, Akshaan Raj, Prasanna Gadbail, Ram Kumar Solanki, and Dr Pawan R. Bhaladhare. "Efficient Approach for Detecting Cardiovascular Disease Using Machine Learning." *Int. J. of Aquatic Science* 14, no. 1 (2023): 308-321.

[13] Al Ahdal, Ahmed, Manik Rakhra, Rahul R. Rajendran, Farrukh Arslan, Moaiad Ahmad Khder, Binit Patel, Balaji Ramkumar Rajagopal, and Rituraj Jain. "Monitoring cardiovascular problems in heart patients using machine learning." *Journal of healthcare engineering* 2023 (2023).

[14] Arsalan Khan, Moiz Qureshi, Muhammad Daniyal, Kassim Tawiah, "A Novel Study on Machine Learning Algorithm-Based Cardiovascular Disease Prediction", *Health & Social Care in the Community*, vol. 2023, Article ID 1406060, 10 pages, 2023.
https://doi.org/10.1155/2023/1406060

[15] Mahboobeh Jafari, Afshin Shoeibi, Marjane Khodatars, Navid Ghassemi, Parisa Moridian, Roohallah Alizadehsani, Abbas Khosravi, Sai Ho Ling, Niloufar Delfan, Yu-Dong Zhang, Shui-Hua Wang, Juan M. Gorriz, Hamid Alinejad-Rokny, U. Rajendra Acharya,
Automated diagnosis of cardiovascular diseases from cardiac magnetic resonance imaging using deep learning models: A review, Computers in Biology and Medicine, Volume 160, 2023, 106998, ISSN 0010-4825, https://doi.org/10.1016/j.compbiomed.2023.106998.
https://www.sciencedirect.com/science/article/pii/S0010482523004638